\documentclass[letterpaper, 10 pt, conference]{ieeeconf}  

\IEEEoverridecommandlockouts                              

\usepackage{multirow}
\usepackage{graphicx}
\usepackage[utf8]{inputenc}
\usepackage[T1]{fontenc}
\usepackage{xcolor}
\usepackage{comment}
\usepackage{amssymb}
\usepackage{amsmath}
\usepackage{booktabs, multirow}
\usepackage{subcaption}
\usepackage{pifont}
\usepackage{array, makecell}
\usepackage{url}
\usepackage{mathrsfs}
\usepackage[colorlinks=true,linkcolor=cyan,urlcolor=red]{hyperref}
\usepackage{cite}
\hypersetup{
    colorlinks=true,
    linkcolor=black,
    filecolor=magenta,
    citecolor=black,
    urlcolor=black,
    }

\title{\LARGE \bf
Domain Adversarial Training for Mitigating Gender Bias in Speech-based Mental Health Detection
}

\author{June-Woo Kim$^{1,2}$ Haram Yoon$^{3}$ Wonkyo Oh$^{3}$ Dawoon Jung$^{3}$ \\ 
Sung-Hoon Yoon$^{1,3}$ Dae-Jin Kim$^{1}$ Dong-Ho Lee$^{3}$ Sang-Yeol Lee$^{1,3}$ Chan-Mo Yang$^{1,3}$$^{\dagger}$
\\ kaen2891@gmail.com \quad ychanmo@wku.ac.kr
\thanks{\line(1,0){150}}
\thanks{$^{\dagger}$Corresponding author}  
\thanks{This research was supported by a grant of the Korea Health Technology R\&D Project through the Korea Health Industry Development Institute (KHIDI), funded by the Ministry of Health \& Welfare, Republic of Korea (grant number: HI22C1962), and by a grant of the Mental Health related Social Problem Solving Project, Ministry of Health \& Welfare, Republic of Korea(grant number: RS-2024-00403474), and by the Bio Industry Technology Development Program funded by the Ministry of Trade, Industry \& Energy (MOTIE, Korea) (Project Number: RS-2024-00431485), and by Brian Impact Foundation, a non-profit organization dedicated to the advancement of science and technology for all.}
\thanks{$^{1}$Department of Psychiatry, Wonkwang University Hospital, Republic of Korea $^{2}$RSC LAB, MODULABS, Republic of Korea $^{3}$Department of Psychiatry, School of Medicine, Wonkwang University, Republic of Korea}
}

\begin{document}

\newcommand{\jw}[1]{{\color{orange} #1}}

\maketitle
\thispagestyle{empty}
\pagestyle{empty}

\begin{abstract}
Speech-based AI models are emerging as powerful tools for detecting depression and the presence of Post-traumatic stress disorder (PTSD), offering a non-invasive and cost-effective way to assess mental health. However, these models often struggle with gender bias, which can lead to unfair and inaccurate predictions. In this study, our study addresses this issue by introducing a domain adversarial training approach that explicitly considers gender differences in speech-based depression and PTSD detection. Specifically, we treat different genders as distinct domains and integrate this information into a pretrained speech foundation model. We then validate its effectiveness on the E-DAIC dataset to assess its impact on performance. Experimental results show that our method notably improves detection performance, increasing the F1-score by up to 13.29 percentage points compared to the baseline. This highlights the importance of addressing demographic disparities in AI-driven mental health assessment.

\indent \textit{Clinical relevance}— Our proposed domain adversarial method can improve the fairness of AI models by mitigating gender bias, thereby improving their clinical applicability and trustworthiness in speech-based mental health detection.

\end{abstract}

\section{INTRODUCTION}
Depression and post-traumatic stress disorder (PTSD) are among the most prevalent and debilitating mental health disorders worldwide, affecting hundreds of millions of individuals~\cite{Christopher2020GlobalBO, Yehuda2015PTSD}.
Both disorders pose significant risks, including increased mortality rates~\cite{Walker2015MortalityIM, Cai2021PrevalenceOS} and long-term socioeconomic burdens~\cite{Davis2022Economic}. Despite their profound impact, diagnosing these conditions remains challenging due to their reliance on subjective clinical assessments rather than objective biomarkers~\cite{Greene2016PTSD, Anderson2024DepressionU}. This subjectivity highlights the need for more reliable and accessible diagnostic tools~\cite{Kapur2012WhyBP}.


One of the considerable approaches for this is speech-based analysis. Unlike costly and time-consuming neuroimaging techniques like EEG and fMRI~\cite{Boby2024EEGdep, Elnaggar2025EEGdep, Kaiser2015LargeScaleND}, speech is a non-invasive and easily accessible modality. 
Research has shown that mental health conditions affect speech patterns, with depression characterized by monotonic prosody and slower speech rates, while PTSD manifests in tense voice features and reduced vowel space~\cite{Cummins2015, Scherer2013AutomaticBD, Scherer2016SelfReportedSO}.

Speech-based mental health benchmark datasets such as Distress Analysis Interview Corpus (DAIC-WOZ)~\cite{Gratch2014DAIC} and extended DAIC (E-DAIC)~\cite{ringeval2019avec} have been widely adopted in research studies~\cite{Ma2016DepAudioNetAE, Gupta2024RADIANCERA}. 
Recent advancements in pretrained speech models have significantly improved classification performance~\cite{Huang2024DepressionRU, Zhang2024ImprovingSD}. However, because these models depend on vocal features, concerns have been raised that their performance might be overestimated due to gender bias~\cite{Bailey2020GenderBI}.
Differences in voice characteristics between male and female speakers can lead to biases in model predictions, which ultimately affect the generalizability of depression classification models.

Mitigating such various demographic biases is critical for developing robust and fair models, as gender disparities in machine learning-based medical diagnostics have been observed across various domains. For instance, Bailey \textit{et al.}~\cite{Bailey2020GenderBI} emphasized that gender-related biases can be performance degraded, therefore they proposed data resampling strategies aimed at balancing gender ratios. 
Other works~\cite{kim2024stethoscope, Huang2023lungsoundDA} tackled electronic stethoscope bias (i.e., recording device for lung sound) by employing domain adversarial training~\cite{ganin2016domain} approach, leading to significant improvements in respiratory sound classification performance. 
In this context, domain adaptation techniques extract domain-invariant features, helping to mitigate diagnostic errors caused by demographic differences in data distribution.
However, no prior studies have specifically explored domain adaptation as a solution for mitigating gender bias in speech-based mental health detection. There is a call for bridging this gap by addressing these issues and enhancing the fairness of AI-driven mental health assessments.

In this study, we leverage a large-scale speech foundation model to classify depression and PTSD while addressing gender bias issues through the domain adversarial training approach. We conduct our experiments using the E-DAIC~\cite{ringeval2019avec} dataset, applying gender domain adaptation to improve model generalizability. Our results demonstrate that incorporating gender adaptation improves the F1-score by up to 13.29 percentage points compared to conventional methods. We believe that our proposed approach would contribute to future medical AI applications, particularly in handling demographic-related bias in mental health diagnostics.
\section{RELATED WORKS}
Recent advancements in deep learning have rapidly improved the detection performance of depression and the presence of PTSD. In one of the earliest works in this area, Ma \textit{et al.}~\cite{Ma2016DepAudioNetAE} introduced DepAudioNet, which integrated CNN and LSTM architectures for training the model. They also introduced random sampling to mitigate class imbalance issues on the DAIC-WOZ dataset~\cite{Gratch2014DAIC}. Recent studies have increasingly focused on leveraging pretrained speech foundation models to drive further advancements in this research field. Huang \textit{et al.}~\cite{Huang2024DepressionRU} utilized wav2vec 2.0 as a feature extractor, while Zhang \textit{et al.}~\cite{Zhang2024ImprovingSD} further advanced this approach by combining wav2vec 2.0 with 1D-CNN and attention mechanisms. They have specifically mitigated the challenges of limited resources through transfer learning techniques.

Despite these promising advancements, concerns regarding gender bias in the DAIC-WOZ~\cite{Gratch2014DAIC} dataset have been raised. Bailey \textit{et al.}~\cite{Bailey2020GenderBI} highlighted that gender-related biases could lead to inflated performance metrics, potentially overestimating model effectiveness. To mitigate this bias, they proposed data redistribution strategies to balance gender ratios within the dataset. While previous efforts primarily focused on data sampling techniques to address imbalanced data or gender bias issues, our study extends this discussion by exploring domain adaptation as a more structured approach for handling demographic imbalances in medical AI. 
\section{PRELIMINARIES}
\subsection{Dataset Description}
We employed the Extended Distress Analysis Interview Corpus (E-DAIC)~\cite{ringeval2019avec} dataset, which is an expansion of the DAIC-WOZ~\cite{Gratch2014DAIC} dataset, designed to support research in automatic mental health assessment, specifically for detecting depression and the presence of PTSD. The E-DAIC dataset includes audio-visual recordings and transcripts of semi-structured clinical interviews between human participants and a virtual agent. These interviews contain a mix of predefined and dynamically adjusted questions. The dataset consists of 275 interview recordings, totaling approximately 73.6 hours, with an average interview length of 16 minutes, and are officially divided into training (165 samples), development (56 samples), and test (56 samples) sets, respectively. Each sample is annotated based on depression status--either depressed (\textbf{D}) or non-depressed (\textbf{ND})--and PTSD status, classified as either PTSD-positive (\textbf{P}) or PTSD-negative (\textbf{NP}).

\begin{table}[t!]
    \centering
    \caption{E-DAIC Dataset distribution for various tasks}
    \renewcommand{\arraystretch}{1.3} 
    \setlength{\tabcolsep}{10pt} 
    \resizebox{\linewidth}{!}{
    
    \begin{tabular}{llcc}
        \toprule
        \textbf{Task} & \textbf{Label} & \textbf{Training} & \textbf{Test} \\
        \midrule
        \multirow{2}{*}{Depression} 
         & Depressed       & 1,658 (23.4\%)  & 624 (30.2\%)  \\
         & Non-depressed   & 5,415 (76.6\%)  & 1,443 (69.8\%) \\
        \midrule
        \multirow{2}{*}{PTSD} 
         & PTSD            & 2,306 (32.6\%)  & 1,218 (58.9\%)  \\
         & Non-PTSD        & 4,767 (67.4\%)  & 849 (41.1\%) \\
        \midrule
        \multirow{2}{*}{Gender} 
         & Male            & 3,284 (46.4\%)  & 1,638 (79.2\%)  \\
         & Female          & 3,789 (53.6\%)  & 429 (20.8\%) \\
        \midrule
        \multirow{4}{*}{Gender-Depression} 
         & Male (D)        & 719 (18.22\%)   & 308 (18.80\%)  \\
         & Male (ND)       & 3,227 (81.78\%) & 1,330 (81.20\%)  \\
         & Female (D)      & 939 (30.03\%)   & 316 (73.66\%)  \\
         & Female (ND)     & 2,188 (69.97\%) & 113 (26.34\%) \\
        \midrule
        \multirow{4}{*}{Gender-PTSD} 
         & Male (P)        & 1,066 (27.01\%) & 580 (35.41\%) \\
         & Male (NP)       & 2,880 (72.99\%) & 1,058 (64.59\%) \\
         & Female (P)      & 1,240 (39.65\%) & 269 (62.70\%) \\
         & Female (NP)     & 1,887 (60.35\%) & 160 (37.30\%) \\
        \bottomrule
    \end{tabular}
    }
    
    \label{tab:tab1}
\end{table}
\subsection{Preprocessing Details}
Due to the considerable length of each interview sample in the E-DAIC dataset, further data preprocessing based on participants' speech was required. To this end, we implemented a 10-second segmentation approach that divides continuous speech into smaller, more manageable units. Specifically, participant speech samples were sequentially concatenated until they reached a predefined threshold of either 10 seconds in duration or five samples. If adding speech exceeded this limit, the accumulated speech was stored as a separate segment, and a new segment was initiated. This method ensured that nearly all individual audio files did not exceed 10 seconds, except in cases where a single sample alone surpassed this duration.

Additionally, the E-DAIC dataset contains a substantial amount of background noise, necessitating additional data filtering. In the initial portion of the recordings, we identified and removed five speech samples that appeared to originate from experiment administrators rather than participants. Furthermore, the final two labeled segments in every sample contained various background noises and annotation errors, leading to their exclusion from the dataset.

Throughout this segmentation process, relevant metadata, such as participant identifiers, depressed and PTSD labels, were carefully maintained and aligned with their corresponding audio segments. As a result, we obtained 7,073 segments for training and 2,067 for testing, as described in Table~\ref{tab:tab1}.

\begin{figure*}[ht!]
    \centering
    \includegraphics[width=0.85\linewidth]{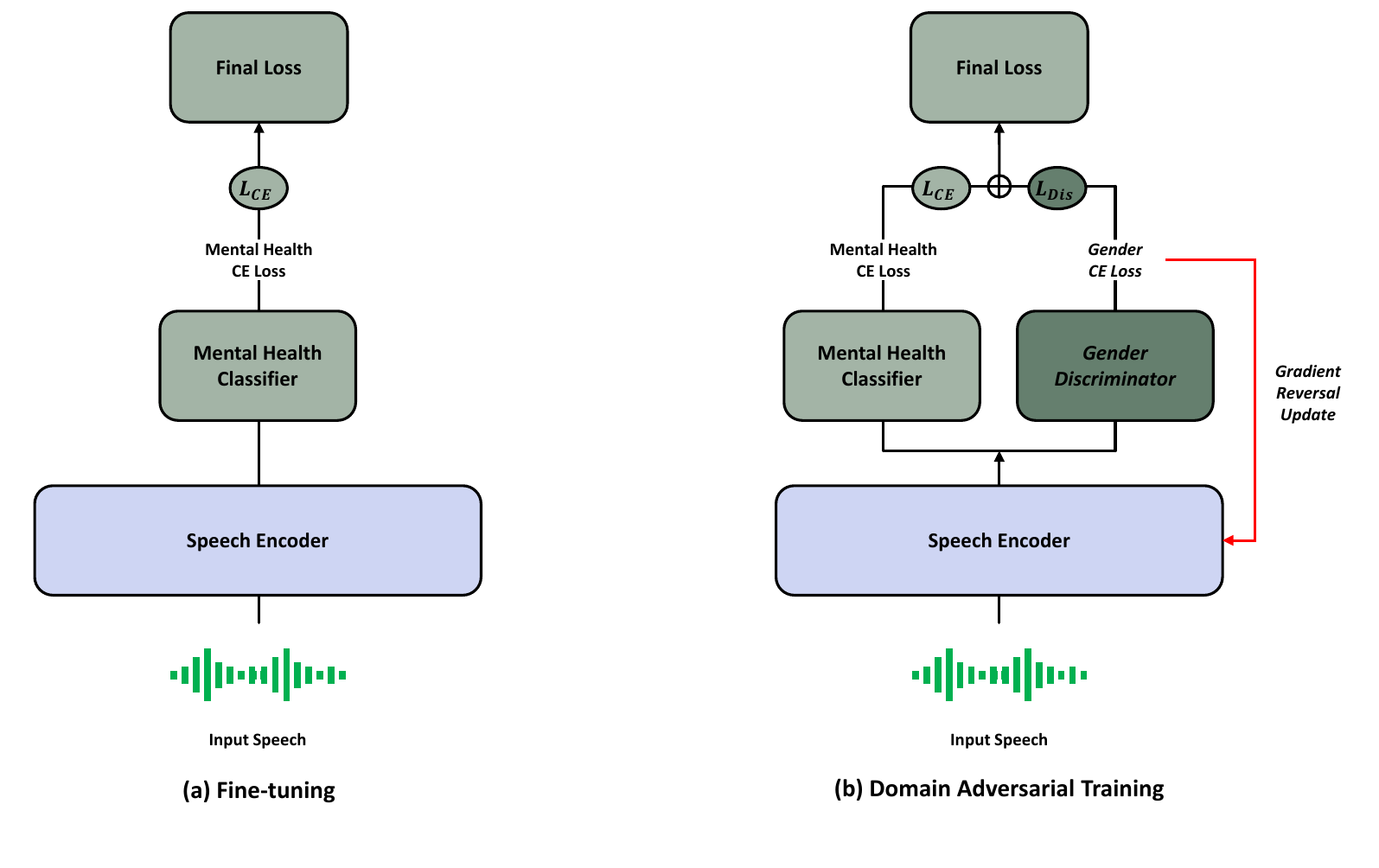}
    \caption{Illustration of fine-tuning and the proposed domain adversarial training architectures designed to mitigate gender bias in speech-based mental health detection.}
    \label{fig:fig1}
\end{figure*}

\subsection{Training Details}
We fine-tuned the pretrained speech foundation models for 50 epochs, leveraging the Adam optimizer with a learning rate of 5e--5, a cosine learning rate schedule, and a batch size of 8. To enhance training stability, we incorporated a momentum update with a coefficient of 0.5 across all learnable parameters as shown in~\cite{bae23b_interspeech}. To mitigate the class imbalance issue, we employed a weighted cross-entropy loss~\cite{bae23b_interspeech}, where class weights were assigned inversely proportional to the number of samples in that class, ensuring a more balanced learning process. For domain adversarial training, we initialized the domain adaptation parameter $\lambda=0.0096$ and progressively increased it to 1.0 throughout 200 training steps.

\section{METHODS}

\subsection{Speech Foundation Models}
The task of mental health detection presents a significant challenge due to the limited availability of training samples, which hinders the ability to train large deep neural networks effectively. In general, large deep neural networks require massive, well-curated datasets to prevent performance degradation due to over- and underfitting when the models are trained from scratch. To address this limitation, prior studies~\cite{Zhang2024ImprovingSD, Huang2024DepressionRU} have leveraged pretrained speech foundation models~\cite{baevski2020wav2vec} to mitigate data scarcity of training samples. Consequently, pretraining on large-scale speech datasets would be a pivotal strategy for improving the performance of speech-based mental health detection.

In this study, we employ three state-of-the-art self-supervised pretrained speech foundation models: wav2vec 2.0~\cite{baevski2020wav2vec}, HuBERT~\cite{hsu2021hubert}, and WavLM~\cite{chen2022wavlm}. These models have achieved remarkable performance across various speech-related downstream tasks~\cite{yang2021superb}, such as automatic speech recognition, speaker identification, emotion recognition, and depression detection. Each model is comprised of several transformer-based encoder layers and trained on large-scale unlabeled speech data, allowing them to learn robust speech representations well-suited for fine-tuning (transfer learning) in low-resource tasks. In our experiments, we adopt the \textit{base} version of each model pretrained on LibriSpeech~\cite{panayotov2015librispeech} 960 hours, for both depression and PTSD detection tasks.

\subsection{Fine-tuning}
As illustrated in Figure~\ref{fig:fig1}a, we fine-tune the pretrained speech foundation models by appending a mental health classifier to their output layer. Specifically, the speech foundation models serve as speech encoders, extracting meaningful representations from input speech. The 768-dimensional extracted representations from the encoder are then processed through a classifier head, followed by a ReLU activation function and a two-dimensional softmax classifier, generating binary predictions for depression and PTSD detection tasks. The classifier is randomly initialized and trained from scratch during the fine-tuning process.

\subsection{Domain Adversarial Training}
While pretrained speech foundation models can be beneficial for speech-based mental health detection, however, fundamental acoustic differences exist between male and female speakers such as fundamental frequency, pitch, and monotonic prosody, which can introduce biases in voice-based depression detection models. To mitigate the gender bias in data distribution, we propose a simple yet effective Domain Adversarial Training (DAT) approach, inspired by~\cite{ganin2016domain, kim2023adversarial, kim2024stethoscope}. The proposed DAT consists of two key loss functions with mental health loss $\mathcal{L}_\text{Mental}$ and gender discriminator loss $\mathcal{L}_\text{Dis}$:
\begin{align}\label{eq:sdat1} 
\mathcal{L}_{\text{Mental}} \! = -\frac{1}{N}\sum_{i=1}^n\! \, y_{i}\! \, \log \, \!(\hat{y_{i}}),
\end{align}
\begin{align}\label{eq:sdat2} 
\mathcal{L}_{\text{Dis}} \! = -\frac{1}{N}\sum_{i=1}^n\! \, g_{i}\! \, \log \, \!(\hat{g_{i}}).
\end{align}
where $\mathcal{L}_{\text{Mental}}$ and $\mathcal{L}_{\text{Dis}}$ represent the Cross-Entropy loss for mental health (depression or PTSD) labels $y$ and gender labels $g$, respectively. The predicted probabilities $\hat{y}$ and $\hat{d}$ are obtained by passing the learned representations through the mental health classifier and gender discriminator, as illustrated in Figure~\ref{fig:fig1}b. To ensure that the learned features remain domain-invariant and cannot distinguish between male and female characteristics, the gradients from $\mathcal{L}_\text{Dis}$ are multiplied by a negative constant during the backpropagation process. The final training objective is formulated as:
\begin{equation}
    \mathcal{L}_{\text{Final}} = \mathcal{L}_{\text{Mental}} + \lambda \, \! \mathcal{L}_{\text{Dis}}
\end{equation}
where $\lambda$ is a regularization parameter for domain adaptation, following the formulation in~\cite{ganin2016domain}. Our DAT framework aims to minimize mental health classification error while ensuring that the learned representations remain invariant across different given speech types, thereby mitigating gender bias in the model.
\section{EXPERIMENTS}
\subsection{Results}

Table~\ref{tab:tab2} illustrates the performance comparison of three speech foundation models--wav2vec2~\cite{baevski2020wav2vec}, HuBERT~\cite{hsu2021hubert}, and WavLM~\cite{chen2022wavlm}--on both depression and PTSD detection tasks using E-DAIC~\cite{ringeval2019avec} dataset. We employed the F1-score as the primary performance metric for distinguishing between \textbf{D} vs. \textbf{ND} for depression detection, and \textbf{P} vs. \textbf{NP} for the PTSD detection task. The reported F1 average (F1 Avg) represents the mean F1-score across the two classes, providing a comprehensive metric for assessing model performance.

\subsubsection{Depression Detection}
In the fine-tuning setting, all models exhibited relatively low F1-scores for the \textbf{D} class rather than that of \textbf{ND} due to the imbalanced configuration of the E-DAIC dataset, as described in~\ref{tab:tab1}. Specifically, wav2vec2 achieved 39.27\% for the \textbf{D} class, while HuBERT and WavLM recorded 38.55\% and 34.92\%. The relatively lower performance in this category suggests that all the models struggled to accurately detect depression states, likely due to data imbalanced or inherent bias within the dataset.
When applying the proposed DAT for addressing gender bias, substantial improvements were observed across all speech foundation models. The F1-score for the \textbf{D} class remarkably increased, reaching 50.65\% for wav2vec2, 58.18\% for HuBERT, and 53.03\% for WavLM. These improvements indicate that incorporating our proposed DAT helped improve classification performance, particularly for the \textbf{D} group. The overall F1 Avg also showed consistent improvements, with wav2vec2 improving from 57.19\% to 67.46\%, HuBERT from 54.74\% to 68.03\%, and WavLM from 55.50\% to 67.95\%, respectively. In our findings, HuBERT showcased the best overall performance and exhibited the greatest improvement among the three models, achieving up to a 13.29\% performance gain.
\begin{table}[t!]
    \centering
    \caption{Performance comparison of different models on the E-DAIC dataset for depression and PTSD detection using fine-tuning and proposed gender domain adversarial training.}
    \renewcommand{\arraystretch}{1.2}
    \setlength{\tabcolsep}{8pt}
    \resizebox{\linewidth}{!}{
    \begin{tabular}{l|ccc|ccc}
        \toprule
        & \multicolumn{3}{c}{\textbf{Depression}} & \multicolumn{3}{c}{\textbf{PTSD}} \\
        \cmidrule(lr){2-4} \cmidrule(lr){5-7}
        \textbf{Method} & \textbf{F1 (ND)} & \textbf{F1 (D)} & \textbf{F1 Avg} & \textbf{F1 (NP)} & \textbf{F1 (P)} & \textbf{F1 Avg} \\
        \midrule
        \multicolumn{7}{c}{\textit{\textbf{Fine-tuning}}} \\
        wav2vec2 & 75.10 & 39.27 & 57.19 & 72.35 & 46.19 & 59.27 \\
        HuBERT & 70.92 & 38.55 & 54.74 & 66.29 & 50.51 & 58.40 \\
        WavLM & 76.08 & 34.92 & 55.50 & 74.67 & 39.51 & 58.09 \\
        \midrule
        \multicolumn{7}{c}{\textit{\textbf{Domain Adversarial Training}}} \\
        wav2vec2 & 84.27 & 50.65 & 67.46 & 71.78 & 49.76 & 60.77 \\
        HuBERT & 77.88 & 58.18 & 68.03 & 73.41 & 54.98 & 64.20 \\
        WavLM & 82.87 & 53.03 & 67.95 & 64.61 & 54.11 & 59.36 \\
        \bottomrule
    \end{tabular}}
    
    \label{tab:tab2}
\end{table}

\subsubsection{PTSD Detection}
For the PTSD detection task, fine-tuning led to moderate F1-scores across both classes. However, as shown in the depression detection task, the \textbf{P} class exhibited relatively lower F1-scores compared to the \textbf{NP} class. For instance, wav2vec2 yielded an F1-score of 46.19\% for the \textbf{P} class, while HuBERT and WavLM reported 50.51\% and 39.51\% respectively. When DAT was applied, the performance improved in all cases. The F1-score for the \textbf{P} class increased to 49.76\%, 54.98\%, and 54.11\% for wav2vec2, HuBERT, and WavLM respectively, and the overall F1 Avg improved across all models. Notably, HuBRRT achieved the highest overall performance, improving from 58.40\% to 64.20\%, demonstrating its strong adaptability to domain-invariant training.



\subsection{Impact of DAT for addressing gender bias in depression task}
To assess the impact of our proposed method on gender-specific depression classification, we present the sub-divided results from Table~\ref{tab:tab2} in Table~\ref{tab:tab3}, providing a detailed performance on comparison across male and female speakers. The F1 Gender Avg column represents the mean F1-score averaged across both gender groups.

\begin{table}[ht!]
    \centering
    \caption{Gender-specific performance comparison of different models for depression detection by using fine-tuning and the proposed gender domain adversarial training.}
    \renewcommand{\arraystretch}{1.2}
    \setlength{\tabcolsep}{8pt}
    \resizebox{\linewidth}{!}{
    \begin{tabular}{lccccccc}
        \toprule
        \multirow{2}{*}{\textbf{Method}} & \multicolumn{3}{c}{\textbf{Male}} & \multicolumn{3}{c}{\textbf{Female}} & \multirow{2}{*}{\textbf{F1 Gender Avg}} \\
        \cmidrule(lr){2-4} \cmidrule(lr){5-7}
         & \textbf{F1 (ND)} & \textbf{F1 (D)} & \textbf{F1 Avg} & \textbf{F1 (ND)} & \textbf{F1 (D)} & \textbf{F1 Avg} & \\
        \midrule
        \multicolumn{8}{c}{\textit{\textbf{Fine-tuning}}} \\
        wav2vec2 & 79.32 & 22.35 & 50.84 & 43.48 & 61.99 & 52.73 & 51.79 \\
        HuBERT & 76.58 & 31.86 & 54.22 & 35.32 & 47.36 & 41.34 & 47.78 \\
        WavLM & 82.28 & 26.75 & 54.52 & 40.09 & 35.75 & 37.92 & 46.22 \\
        \midrule
        \multicolumn{8}{c}{\textit{\textbf{Domain Adversarial Training}}} \\
        wav2vec2 & 89.24 & 25.65 & 57.45 & 47.41 & 63.84 & 55.63 & 56.54 \\
        HuBERT & 81.38 & 33.68 & 57.53 & 44.16 & 48.63 & 46.40 & 51.97 \\
        WavLM & 87.82 & 28.43 & 58.13 & 42.14 & 40.11 & 41.13 & 49.63 \\
        \bottomrule
    \end{tabular}}
    \label{tab:tab3}
\end{table}

\subsubsection{Depression Results on Fine-tuning}
We observed that notable gender disparities remained across all three models. In particular, male speakers generally exhibited higher F1-scores for \textbf{ND} classification compared to female speakers. For instance, wav2vec2 achieved 79.32\% for \textbf{ND} in males, while the corresponding value for females was considerably lower at 43.48\%. This phenomenon was also found in HuBERT and WavLM. Conversely, depression classification performance showed the opposite trend, with female speakers achieving higher F1-scores than males. wav2vec2 recorded the \textbf{D} of 22.35\% for males, whereas the females reached 61.99\%, indicating a significant imbalance. similar patterns were observed for HuBERT (33.68\% for males vs. 47.86\% for females) and WavLM (28.43\% for males vs. 35.75\% for females). These findings highlight a gender-driven performance gap, where models were better at detecting depression in female speakers but more reliable in classifying \textbf{ND} cases for females.

\subsubsection{Depression Results on Proposed DAT}
Applying our DAT method led to notable performance improvements across both male and female speakers, remarkably reducing the gender-based performance gap. For male speakers, the F1 Avg scores increased from 50.84\% to 57.45\% for wav2vec 2.0, from 54.22\% to 57.53\% for HuBERT, and from 54.52\% to 58.13\% for WavLM. Similarity for the female speakers, the F1 Avg scores improved rising from 52.73\% to 55.63\% for wav2vec 2.0, and the HuBERT and WavLM saw corresponding increases from 41.34\% to 46.40\% and 37.92\% to 41.13\%, respectively. These improvements suggest that DAT enables the models to generalize more effectively across genders, mitigating the performance disparities observed in fine-tuning approaches. The most compelling evidence of this improvement is reflected in the F1 Gender Avg scores, which increased across all models.

\subsection{Impact of DAT for addressing gender bias in PTSD task}

\begin{table}[ht!]
    \centering
    \caption{Gender-specific performance comparison of different models for PTSD detection by using fine-tuning and the proposed gender domain adversarial training.}
    \renewcommand{\arraystretch}{1.2}
    \setlength{\tabcolsep}{8pt}
    \resizebox{\linewidth}{!}{
    \begin{tabular}{lccccccc}
        \toprule
        \multirow{2}{*}{\textbf{Method}} & \multicolumn{3}{c}{\textbf{Male}} & \multicolumn{3}{c}{\textbf{Female}} & \multirow{2}{*}{\textbf{F1 Gender Avg}} \\
        \cmidrule(lr){2-4} \cmidrule(lr){5-7}
         & \textbf{F1 (ND)} & \textbf{F1 (D)} & \textbf{F1 Avg} & \textbf{F1 (NP)} & \textbf{F1 (P)} & \textbf{F1 Avg} & \\
        \midrule
        \multicolumn{8}{c}{\textit{\textbf{Fine-tuning}}} \\
        wav2vec2 & 76.03 & 48.46 & 62.25 & 56.05 & 39.78 & 47.91 & 55.08 \\
        HuBERT & 70.34 & 55.88 & 63.11 & 50.4 & 30.73 & 40.56 & 51.84 \\
        WavLM & 70.73 & 49.5 & 60.12 & 48.25 & 10.79 & 29.52 & 44.82 \\
        \midrule
        \multicolumn{8}{c}{\textit{\textbf{Domain Adversarial Training}}} \\
        wav2vec2 & 77.01 & 50.63 & 63.85 & 54.93 & 47.77 & 51.35 & 57.60 \\
        HuBERT & 77.79 & 59.22 & 68.51 & 53.97 & 42.11 & 48.04 & 58.28 \\
        WavLM & 70.25 & 50.79 & 60.52 & 55.22 & 41.85 & 48.54 & 54.53 \\
        \bottomrule
    \end{tabular}}
    \label{tab:tab4}
\end{table}

Table~\ref{tab:tab4} illustrates a gender-specific performance comparison for PTSD detection from three models under both fine-tuning and our proposed DAT approach. 
\subsubsection{PTSD Results on Fine-tuning}
Under the fine-tuning baseline, for male speakers, models exhibited relatively higher F1-scores across both \textbf{NP} and \textbf{P} categories compared to female speakers. Specifically, wav2vec2, HuBERT, and WavLM for male speakers achieved the \textbf{F1 Avg} scores of 62.25\%, 63.11\%, and 60.12\%, while the comparisons on female speakers achieved 47.91\%, 40.56\%, and 29.52\%, respectively. Specifically, the WavLM results on female speakers obtained 10.79\% of \textbf{P} scores, indicating that the fine-tuned models struggled to correctly classify PTSD cases, leading to a substantial gender imbalance.

\subsubsection{PTSD results on Proposed DAT}
For male speakers, PTSD detection performance improved across all models, as described in Table~\ref{tab:tab4}. More importantly, female speaker performance improved notably, the \textbf{P} scores from wav2vec2, HuBERT, and WavLM increased from 39.78\% to 47.77\%, 30.73\% to 42.11\%, and 10.78\% to 41.85\%, respectively, This suggests that the proposed DAT successfully addressed the gender imbalance issue by enable the models to generalize more effectively across both gender groups. Furthermore, \textbf{F1 Gender Avg} scores also showcased considerable improvements, rising from 55.08\% to 57.60\% for wave2vec2, 51.84\% to 58.28\% for HuBERT, and 44.82\% to 54.53\% for WavLM. These improvements confirm that our proposed method boosts model robustness and fairness, ensuring that both male and female speakers receive more balanced PTSD predictions.

\subsection{Qualitative Analysis}

\begin{figure}[!t]
    \vspace{-1mm}
    \centering
    \begin{subfigure}{.5\linewidth}
      \centering
      \includegraphics[width=1.0\linewidth]{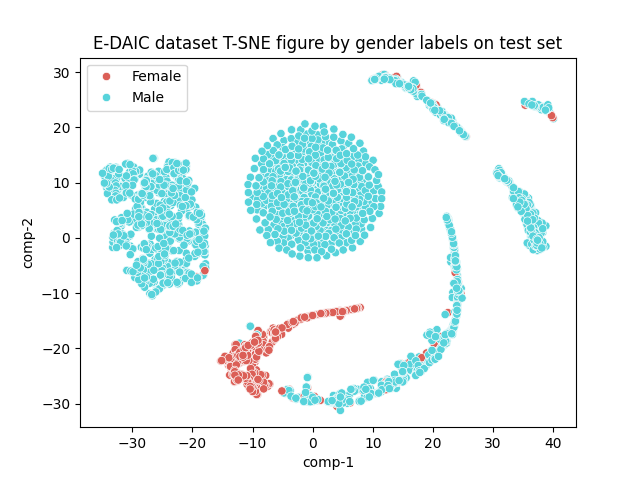}
      \caption{Fine-tuning.}
      \label{fig:sfig1}
    \end{subfigure}%
    \begin{subfigure}{.5\linewidth}
      \centering
      \includegraphics[width=1.0\linewidth]{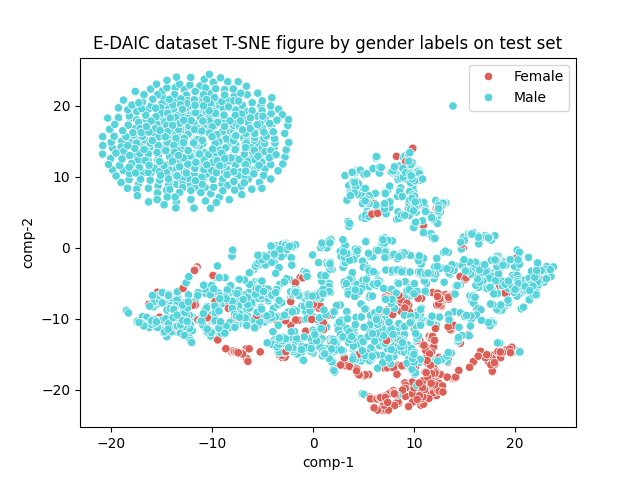}
      \caption{Proposed DAT.}
      \label{fig:sfig2}
    \end{subfigure}  
    \caption{T-SNE results of wav2vec2 fine-tuning and proposed DAT on E-DAIC test set for gender labels.}
    \label{fig:t-sne}
    \end{figure}

To analyze the extent to which our proposed DAT mitigates gender-related biases, Figure~\ref{fig:t-sne} presents t-SNE visualizations of the feature representations from the E-DAIC test set, comparing fine-tuning (Figure~\ref{fig:sfig1}) and DAT (Figure~\ref{fig:sfig2}). In the fine-tuning case, the embeddings exhibit strong clustering by gender, indicating that the model inherently captures and encodes gender-specific biases during training. This suggests that the extracted features are highly dependent on gender attributes, which could contribute to biased decision-making in downstream classification tasks.

In contrast, the DAT-based representation (Figure~\ref{fig:sfig2}) illustrates a more intermixed distribution of male and female data points, indicating that gender-related distinctions have been partially mitigated. This suggests that the proposed DAT method helps the model learn more generalized and unbiased represenations, reducing the extent of gender bias. However, despite these improvements, some degree of gender clustering remains (top-left region in Figure~\ref{fig:sfig2}), suggesting that while DAT reduces gender dependency, it does not completely eliminate it. Future work could explore more advanced debiasing techniques to further improve fairness in speech-based mental health assessments.
\section{CONCLUSION}
This work investigated gender-based performance disparities in speech foundation models for depression and PTSD detection. Through extensive empirical analysis of the E-DAIC dataset, we demonstrated that conventional fine-tuning approaches lead to significant gender-based performance gaps. To address this challenge, we proposed a DAT approach that explicitly considers gender-specific characteristics. Our approach achieved substantial improvements in cross-gender generalization, with particularly strong gains in depression detection performance. Experimental results validate that DAT successfully mitigates gender-based performance disparities while maintaining high F1-scores.

This study highlights the importance of bias-aware learning in medical AI. Future work should be considered for other demographic factors (e.g., age, accent variations), larger datasets, and leveraging LLM to further improve fairness and clinical applicability in AI-driven mental health assessment.



\bibliography{ref}
\bibliographystyle{IEEEtran}

\end{document}